%% file: main.tex
\documentclass[runningheads]{llncs}

\usepackage{graphicx}
\usepackage{comment}
\usepackage{amsmath,amssymb}
\usepackage{color}
\usepackage{url}
\usepackage{hyperref}
\usepackage{bbm}
\usepackage{makecell}
\usepackage{subcaption}
\usepackage[numbers,sort&compress]{natbib}
\makeatletter
\renewcommand\@biblabel[1]{#1.}

\makeatother

\usepackage[capitalize]{cleveref}

\usepackage{tikz}
\usetikzlibrary{arrows}
\tikzstyle{arrow}=[draw, -latex]
\usetikzlibrary{positioning,calc}
\usetikzlibrary{shapes.misc}
\usetikzlibrary{decorations.pathreplacing} 
\usetikzlibrary{decorations.markings}
\usetikzlibrary{fit}
\usetikzlibrary{shapes.callouts}
\usetikzlibrary{shapes.geometric}
\usetikzlibrary{matrix}
\usepgflibrary{arrows.meta}
\usepackage{tikz-3dplot}
\usepackage{tikzscale}
\usetikzlibrary{backgrounds}

\usepackage{setspace}
\usepackage{transparent}
\usetikzlibrary{3d}

\usepackage{pgfplots}
\pgfplotsset{compat=1.9}
\usepgfplotslibrary{groupplots}

\usepgfplotslibrary{units}
\usepackage{pgfplotstable}
\usepackage{booktabs}
\usepackage{multirow}
\usepackage[draft,nomargin,inline]{fixme}
\fxsetup{theme=color}

\DeclareMathOperator*{\argmin}{arg\,min}

\newif\ifreview
\reviewfalse

\ifreview
	\usepackage{lineno}

	\linenumbers
\fi

\newcommand{\rotateRPY}[3]% roll, pitch, yaw
{   \pgfmathsetmacro{\rollangle}{#1}
    \pgfmathsetmacro{\pitchangle}{#2}
    \pgfmathsetmacro{\yawangle}{#3}

    \pgfmathsetmacro{\newxx}{cos(\yawangle)*cos(\pitchangle)}
    \pgfmathsetmacro{\newxy}{sin(\yawangle)*cos(\pitchangle)}
    \pgfmathsetmacro{\newxz}{-sin(\pitchangle)}
    \path (\newxx,\newxy,\newxz);
    \pgfgetlastxy{\nxx}{\nxy};

    \pgfmathsetmacro{\newyx}{cos(\yawangle)*sin(\pitchangle)*sin(\rollangle)-sin(\yawangle)*cos(\rollangle)}
    \pgfmathsetmacro{\newyy}{sin(\yawangle)*sin(\pitchangle)*sin(\rollangle)+ cos(\yawangle)*cos(\rollangle)}
    \pgfmathsetmacro{\newyz}{cos(\pitchangle)*sin(\rollangle)}
    \path (\newyx,\newyy,\newyz);
    \pgfgetlastxy{\nyx}{\nyy};

    \pgfmathsetmacro{\newzx}{cos(\yawangle)*sin(\pitchangle)*cos(\rollangle)+ sin(\yawangle)*sin(\rollangle)}
    \pgfmathsetmacro{\newzy}{sin(\yawangle)*sin(\pitchangle)*cos(\rollangle)-cos(\yawangle)*sin(\rollangle)}
    \pgfmathsetmacro{\newzz}{cos(\pitchangle)*cos(\rollangle)}
    \path (\newzx,\newzy,\newzz);
    \pgfgetlastxy{\nzx}{\nzy};
}

\tikzset{RPY/.style={x={(\nxx,\nxy)},y={(\nyx,\nyy)},z={(\nzx,\nzy)}}}

\tikzset{cross/.style={cross out, draw=black, minimum size=2*(#1-\pgflinewidth), inner sep=0pt, outer sep=0pt},
cross/.default={1pt}}

\usepackage{array}
\usepackage[abbreviations=true,per-mode=symbol]{siunitx}
\newcolumntype{C}[1]{>{\centering\arraybackslash}p{#1}}

\begin{document}

%%%%%%%%%%%%%%%%%%%%% Add submission id, track, and title. %%%%%%%%%%%%%%%%%%%%%

% Insert your submission number here
\def\SubNumber{107}

% Choose one track by uncommenting one of the following lines  
\def\GCPRTrack{Track: Robot vision}
% \def\GCPRTrack{Track: Computer vision systems and applications}
% \def\GCPRTrack{Track: Pattern recognition in the life and natural sciences}
% \def\GCPRTrack{Track: Photogrammetry and remote sensing}
% \def\GCPRTrack{Track: Robot vision}
% \def\GCPRTrack{Track: DAGM Young Researcher Forum}

% Replace with your title
\title{T6D-Direct: Transformers for Multi-Object\\ 6D Pose Direct Regression}
% You can use \thanks for acknowledgment. Do not add any acknowledgment to the draft 
% version that is used for the review process.  
%\title{Title\thanks{XXX}}

\ifreview
	% ANONYMOUS SUBMISSION FOR REVIEW
	% DO NOT MODIFY these for the draft version that is used for the review process.
	\titlerunning{DAGM GCPR 2021 Submission \SubNumber{}. CONFIDENTIAL REVIEW COPY.}
	\authorrunning{DAGM GCPR 2021 Submission \SubNumber{}. CONFIDENTIAL REVIEW COPY.}
	\author{DAGM GCPR 2021 - \GCPRTrack{}}
	\institute{Paper ID \SubNumber}
\else

	\author{Arash Amini\inst{1} \and Arul Selvam Periyasamy\inst{2} \and Sven Behnke\inst{2}}
	
	\authorrunning{Amini et al.}
	
	\institute{Autonomous Intelligent Systems, University of Bonn \\ Germany \\ \email{amini@uni-bonn.de}\inst{1}, \email{\{periyasa,behnke\}@ais.uni-bonn.de}\inst{2}}
\fi

\maketitle              % typeset the header of the contribution

\begin{abstract}
6D pose estimation is the task of predicting the translation and orientation of objects in a given input image, which is a crucial prerequisite for many robotics and augmented reality applications. Lately, the Transformer Network architecture, equipped with multi-head self-attention mechanism, is emerging to achieve state-of-the-art results in many computer vision tasks. DETR, a Transformer-based model, formulated object detection as a set prediction problem and achieved impressive results without standard components like region of interest pooling, non-maximal suppression, and bounding box proposals. In this work, we propose T6D-Direct, a real-time single-stage direct method with a transformer-based architecture built on DETR to perform 6D multi-object pose direct estimation. We evaluate the performance of our method on the YCB-Video dataset. Our method achieves the fastest inference time, and the pose estimation accuracy is comparable to state-of-the-art methods.

\keywords{pose estimation  \and transformer \and self-attention.}
\end{abstract}

\section{Introduction}
% \textbf{Please add the paper submission id and the track name to the paper.}
6D object pose estimation in clutter is a necessary prerequisite for autonomous robot manipulation tasks and augmented reality.
Given the complex nature of the task, methods for object pose estimation---both traditional and 
modern---are multi-staged~\cite{xiang2017posecnn, hodavn2020bop, schwarz2018fast, Oberweger2018}.
The standard pipeline consists of an object detection and/or instance segmentation, followed by
the region of interest cropping and processing the cropped patch to estimate the 6D pose of an object. 
Convolutional neural networks (CNNs) are the basic building blocks of the deep learning models for computer vision tasks.
CNN's strength lies in the ability to learn local spatial features.
Motivated by the success of deep learning methods for computer vision, in a strive for end-to-end differentiable pipelines, many of the traditional components like
non-maximum suppression (NMS) and region of interest cropping (RoI) have been replaced by their differentiable counterparts~\cite{hosang2017learning, girshick2015fast, ren2015faster}.
Despite these advancements, the pose estimation accuracy still heavily depends on the initial object detection stage.

Recently, Transformer, an architecture based on self-attention mechanism, is achieving state-of-the-art results in many natural language processing
tasks. Transformers are efficient in modeling long-range dependencies in the data, which is also beneficial for many computer vision tasks. Some recent works achieved state-of-the-art results in computer vision tasks using the Transformer architecture to supplement CNNs
or to completely replace CNNs ~\cite{dosovitskiy2021an, touvron2020training, carion2020end, zhu2020deformable, wang2020axial, khan2021transformers}.

Carion et al. introduced DETR~\cite{carion2020end}, an object detection pipeline using Transformer in combination with a CNN backbone model and achieved impressive
results. DETR is a simple architecture without any handcrafted procedures like NMS and anchor generation. It formulates object detection as a
set prediction problem and uses bipartite matching and Hungarian loss to implement an end-to-end differentiable pipeline for object detection.

In this paper, we present T6D-Direct, an extension to the DETR architecture to perform multi-object 6D pose direct regression in real-time. T6D-Direct enables truly end-to-end pipeline for 6D object pose estimation where the accuracy of the pose estimation is not reliant on object detection and the subsequent cropping. In contrast to the standard methods of 6D object pose estimation that are multi-staged, our method is direct single-stage and estimates the pose of all the objects in a given image in one forward pass. In short, our contributions include:

\begin{enumerate}
\item An elegant real-time end-to-end differentiable architecture for multi-object 6D pose direct regression.
\item Evaluation of different design choices for implementing multi-object 6D pose direct regression as a set prediction problem.

\end{enumerate}

\section{Related Work}

In this section, we review the state-of-the-art methods for 6D object pose estimation and DETR, 
the transformer architecture our proposed method is based on, in detail. 
\subsection{Pose Estimation}
Like most other computer vision tasks, the state-of-the-art methods for 6D object pose estimation from RGB images are predominantly convolutional neural network (CNN)-based. The standard CNN architectures for object pose estimation are multi-staged. The first stage is object detection and/or instance segmentation.
In the second stage, using the object bounding boxes, predicted an image patch containing the target object, is extracted and the 6D pose of the object is estimated. The common methods for object pose estimation can be broadly classified into three categories: direct, indirect, and refinement-based.

Direct methods regress for the translation and orientation components of the object pose directly from the RGB images~\citep{xiang2017posecnn,periyasamy2018pose, Hu2020CVPR}. ~\citet{kehl2017ssd, Sundermeyer_2018_ECCV} discretized the orientation component of the 6D pose and performed classification instead of regression.

Indirect approaches aim to recover the 6D pose from the 2D-3D correspondences using the P\textit{n}P algorithm, where P\textit{n}P is often used in combination with the RANSAC algorithm to increase the robustness against outliers in correspondence prediction ~\citep{rad2017bb8,tekin2018real,hu2019segmentation,peng2019pvnet}. Although indirect methods outperform direct methods in the recent benchmarks~\citep{hodavn2020bop}, indirect models are significantly larger, and the model size grows exponentially with the number of objects. One common solution to keep the model size small is to train one lighter model for each object. This approach, however, introduces significant overhead for many real-world applications. Direct models, on the other hand, are lighter, and their end-to-end differentiable nature is desirable in many applications~\citep{wang2020self6d}.
~\citet{Wang_2021_GDRN, Li2019ICCV} unified direct regression and dense estimation methods by introducing a learnable P\textit{n}P module. 

Refinement-based methods formulate 6D pose estimation as an iterative refinement 
problem where in each step given the observed image and the rendered image according to the current pose estimate, 
the model predicts a pose update that aligns the observed and the rendered image better. The process is repeated until the estimated pose update is negligibly small. 
Refinement-based methods are orthogonal to the direct and indirect methods and are often used in combination with these methods~\cite{labbe2020, manhardt2018deep, Shao_2020_CVPR, periyasamy2019refining, li2018deepim}, i.e., direct or indirect methods produce initial pose estimate and the refinement-based methods are used to refine the initial pose estimate to predict the final accurate pose estimate.

% \subsection{Transformers for Computer Vision Tasks}

%Transformer, a neural network architecture based on self-attention mechanism, introduced by~\citet{vaswani2017attention} for modeling long-term dependencies quickly became the foundation for 
%many state-of-the-art natural language processing models. Recently, architectures based on Transformer are starting to achieve state-of-the-art results in many computer vision tasks as well.
%Notably, image recognition~\cite{dosovitskiy2021an, touvron2020training}, object detection~\cite{carion2020end, zhu2020deformable}, image segmentation~\cite{carion2020end, wang2020axial}, 
%image generation~\cite{parmar2018image, chen2020generative, esser2020taming}.

%In many cases, Transformers are replacing CNNs completely~\cite{dosovitskiy2021an, touvron2020training} and in other cases,
%Transformers are used in combination with CNNs to replace typical building blocks like non-maximum suppression and ROI pooling~\cite{carion2020end, zhu2020deformable}. 
%For a detailed survey of Transformers for computer vision tasks, we refer the readers to the survey by~\citet{khan2021transformers}. 

\subsection{DETR}
\citet{carion2020end} introduced DETR, an end-to-end differentiable object detection model
using the Transformer architecture. They formulated object detection, the problem of 
estimating the bounding boxes and class label probabilities, as a set prediction problem. Given an RGB input image, the DETR model outputs 
a set of tuples with fixed cardinality. Each tuple consists of the bounding box and class label probability of an object. To allow an output set with
a fixed cardinality, a larger cardinality is chosen, and a special class id \text{\O} is used for padding the rest of the tuples in addition to the actual object detections.
The tuples in the predicted set and the ground truth target set are matched by bipartite matching using the Hungarian algorithm. The DETR model achieved competitive results on the COCO dataset~\citep{lin2014microsoft} compared to standard CNN-based architectures. 

\section{Method}

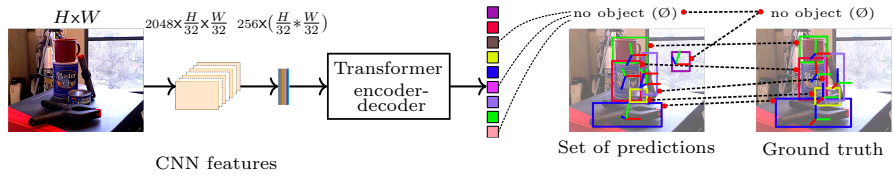
\begin{figure}[t]
	\centering
	  \resizebox{.99\linewidth}{!}{\input{pipeline.tikz}}
	  \caption{T6D-Direct overview. Given an RGB image, we use a CNN backbone to extract lower-resolution image features and flatten them to
	  create feature vectors suitable for a standard Transformer model. The Transformer model generates a set of predictions with a fixed cardinality $N$. 
	  To facilitate the prediction of a varying number of objects
	  in an image, we choose $N$ to be much larger than the expected number of objects in an image and pad the rest of the tuples in the set with \text{\O} object
	  predictions. We perform bipartite matching between the predicted and ground truth sets to find the matching pairs and train the pipeline to minimize
	  the Hungarian loss between the matched pairs.}
	  \label{fig:pipeline}
\end{figure}

In this section, we describe our approach of formulating 6D object pose estimation as a set prediction problem and describe the extensions we made to the DETR model
and the bipartite matching process to enable the prediction of a set of tuples of bounding boxes, class label probabilities, and 6D object poses. \cref{fig:pipeline}
provides an overview of the proposed T6D-Direct model.

\subsection{Pose Estimation as Set Prediction}
Inspired by the DETR model, we formulate 6D object pose direct regression as a set prediction problem. We call our method T6D-Direct. In the following sections, we describe the individual components of the T6D-Direct model in detail.
\subsubsection{Set Prediction}
 Given an RGB input image, our model generates a set of tuples. Each tuple consists of a bounding box, represented as center coordinates, height and width, class label probabilities,
 translation and orientation components of the 6D object pose. The height and width of the bounding boxes are proportional to the size of the image. For the orientation component, we opt for
 the 6D continuous representation as shown to yield the best performance in practice~\citep{zhou2019continuity}.
 To facilitate the 6D pose prediction of a varying number of objects in an image, we fix the cardinality of the predicted set to $N$, which is a hyperparameter, and we chose it to be larger than the expected maximum
 number of objects in the image. In this way, the network has enough options to embed each object freely. The T6D-Direct model is trained to predict the tuples corresponding to the objects in the image and predict \text{\O} class to pad the rest of the tuples in the fixed size set.
\subsubsection{Bipartite Matching}
Given $n$ ground truth objects ${y}_1, {y}_2, ..., {y}_n$, we pad \text{\O} objects to create a ground truth set $y$ of cardinality $N$. 
To match the predicted set $\hat{y}$, generated by our T6D-Direct model, with the ground truth set $y$, we perform bipartite matching. Formally, we search for the permutation of elements between the two sets $\sigma \in \mathfrak{S}_{N}$ that minimizes the matching cost:

\begin{equation}
\hat{\sigma} = \argmin_{\sigma \in \mathfrak{S}_{N}} \sum_i^N  \mathcal{L}_{match}(y_i, \hat{y}_{\sigma(i)}),
\end{equation}
where $ \mathcal{L}_{match}(y_i, \hat{y}_{\sigma(i)}) $ is the pair-wise matching cost between the ground truth tuple $y_i$
and the prediction at index $\sigma(i)$. DETR model included bounding boxes $b_i$ and class probabilities $p_i$ in their cost function.
In the case of T6D-Direct model, we have two options for defining $\mathcal{L}_{match}(y_i, \hat{y}_{\sigma(i)})$. One option is to use the same definition used by the
DETR model, i.e., we include only bounding boxes and class probabilities and ignore pose predictions in the matching cost.
We call this variant of matching cost as $\mathcal{L}_{match\_object}$.

\begin{equation}
\mathcal{L}_{match\_object}(y_i, \hat{y}_{\sigma(i)}) = -\mathbbm{1}_{c_i\neq \text{\O}} \hat{p}_{\sigma(i)}(c_i) + \mathbbm{1}_{c_i\neq \text{\O}} \mathcal{L}_{box}(b_i, \hat{b}_{\sigma(i)}).
\end{equation}

The second option is to include the pose predictions in the matching cost as well. We call this variant  $\mathcal{L}_{match\_pose}$.

\begin{multline}\label{match_pose}
\mathcal{L}_{match\_pose}(y_i, \hat{y}_{\sigma(i)}) =  \mathcal{L}_{match\_object}(y_i, \hat{y}_{\sigma(i)}) + \\
\mathcal{L}_{rot}(R_i, \hat{R}_{\sigma(i)}) +
\mathcal{L}_{trans}(t_i, \hat{t}_{\sigma(i)}),
\end{multline}
where $\mathcal{L}_{rot}$ is the angular distance between the ground truth and predicted rotations, and $\mathcal{L}_{trans}$ is the $\ell_2$ loss between the ground truth and estimated translations. We experimented with both variants, and we opted for the former method.
% as this approach slightly obtains better results.

\subsubsection{Hungarian Loss}
After establishing the matching pairs using the bipartite matching, the T6D-Direct model is trained to minimize the \textit{Hungarian loss} between the 
predicted and ground truth target sets consisting of probability loss, bounding box loss, and pose loss:

\begin{multline}\label{hungarian_loss}
\mathcal{L}_{Hungarian}(y, \hat{y}) = \sum_i^N [-\text{log}\hat{p}_{\hat{\sigma}(i)}(c_i) +
			\mathbbm{1}_{c_i\neq \text{\O}} \mathcal{L}_{box}(b_i, \hat{b}_{\hat{\sigma}(i)}) + \\
			\lambda_{pose}\mathbbm{1}_{c_i\neq \text{\O}} \mathcal{L}_{pose}(R_i, t_i, \hat{R}_{\hat{\sigma}(i)}, \hat{t}_{\hat{\sigma}(i)}) ].
\end{multline}

\subsubsection{Class probability loss}
The first component in the \textit{Hungarian loss} is the class probability loss. We use the standard negative log-likelihood loss as the
class probabilities loss function. Additionally, the number of \text{\O} classes in a set is significantly larger than the other object classes. To counter this class imbalance, we weight the log probability loss for the \text{\O} class by a factor of 0.4.

\subsubsection{Bounding box loss}
The second component in the \textit{Hungarian loss} is bounding box loss $\mathcal{L}_{box}(b_i, \hat{b}_{\sigma(i)})$. We use a weighted 
combination of generalized \mbox{IOU~\citep{rezatofighi2019generalized}} and $\ell_1$ loss.

\begin{equation}\label{eqn:box_loss}
\mathcal{L}_{box}(b_i, \hat{b}_{\sigma(i)}) = \alpha \mathcal{L}_{iou}(b_i, \hat{b}_{\sigma(i)}) + \beta || b_i - \hat{b}_{\sigma(i)} ||,
\end{equation}
\begin{equation}
\mathcal{L}_{iou}(b_i, \hat{b}_{\sigma(i)}) = 1 - \left( \frac{|b_i \cap \hat{b}_{\sigma(i)}|}{|b_i \cup \hat{b}_{\sigma(i)}|} - \frac{|B(b_i, \hat{b}_{\sigma(i)}) \setminus b_i \cup \hat{b}_{\sigma(i)} |}{|B(b_i, \hat{b}_{\sigma(i)})|}  \right),
\end{equation}
where $\alpha$, $\beta$ are hyperparameters and  $B(b_i, \hat{b}_{\sigma(i)})$ is the largest box containing both the ground truth $b_i$ and the prediction $\hat{b}_{\sigma(i)}$.
\subsubsection{Pose loss}
\label{sec:poseloss}

The third component of the \textit{Hungarian loss} is the pose loss. Inspired by~\citet{Wang_2021_GDRN}, we use the disentangled loss to individually supervise the translation $t$ and rotation $R$ via employing symmetric aware loss~\citep{xiang2017posecnn} for the rotation, and $\ell_2$ loss for the translation.

\begin{equation}\label{eqn:ploss}
\mathcal{L}_{pose}(R_i, t_i, \hat{R}_{\sigma(i)}, \hat{t}_{\sigma(i)}) = \mathcal{L}_{R}(R_i, \hat{R}_{\sigma(i)}) + || t_i - \hat{t}_{\sigma(i)} ||,
\end{equation}

\begin{equation}\label{eqn:pose_loss}
\mathcal{L}_{R} = \left\{\begin{array}{ll}
\frac{1}{|\mathcal{M}|} \displaystyle\sum_{\text{x}_1 \in \mathcal{M}}  \min_{\text{x}_2 \in \mathcal{M}}|| (R_i\text{x}_1 - \hat{R}_{\sigma(i)} \text{x}_2) || & \text { if symmetric, } \\
\frac{1}{|\mathcal{M}|} \displaystyle\sum_{\text{x} \in \mathcal{M}} || (R_i\text{x} - \hat{R}_{\sigma(i)} \text{x}) || & \text { otherwise, } 
\end{array}\right.
\end{equation}
where $\mathcal{M}$ indicates the set of 3D model points. Here, we subsample 1500 points from provided meshes. $R_i$ is the ground truth rotation and $t_i$ is the ground truth translation. $\hat{R}_{\sigma(i)}$ and $\hat{t}_{\sigma(i)}$ are the predicted rotation and translation, respectively.

\subsection{T6D-Direct architecture}
\begin{figure}
	\centering
	  \resizebox{.99\linewidth}{!}{\input{architecture.tikz}}
	  \caption{T6D-Direct architecture in detail. Flattened positional encoded image features from a backbone model are made available to
	  each layer of the transformer encoder. The output of the encoder is provided as input to the decoder along with positional encoding. 
	  But, unlike the encoder that takes fixed sine positional encoding, we provide learned positional encoding to the decoder.
	  We call these learned positional encoding \textit{object queries}. Each output of the decoder is processed independently in parallel by shared prediction
	  heads to generate a set of $N$ tuples each containing class probabilities, bounding boxes and translation and orientation components of the 6D object pose.
	  Since the cardinality of the set is fixed, after predicting all the objects in the given image, we train the model to predict \text{\O} object for the rest of the
	  tuples.
	  }
	  \label{fig:architecture}
\end{figure}
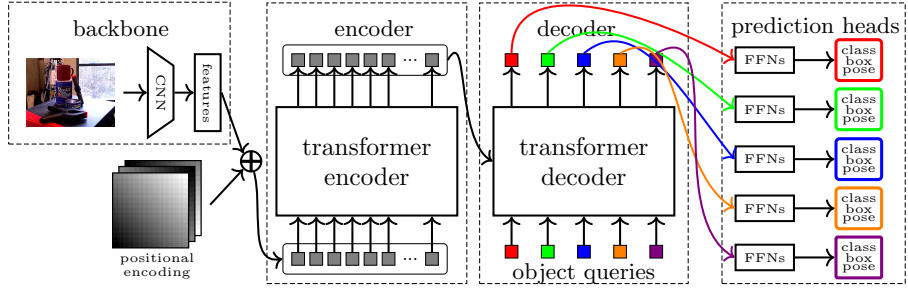

The proposed T6D-Direct architecture for 6D pose estimation is largely based on DETR architecture. 
We use the same backbone CNN (ResNet50), positional encoding, and the transformer encoder and decoder components of the DETR architecture. 
The only major modification is adding feed-forward prediction heads to predict
the translation and rotation components of 6D object poses in addition to the bounding boxes and the class probabilities. We discuss the individual components 
of the T6D-Direct architecture in detail in the following sections.
\subsubsection{CNN feature extraction and positional encoding}
We use ResNet50~\citep{he2016deep} model pretrained on ImageNet~\citep{deng2009imagenet} with frozen batch normalization layers to extract features from the input RGB image. Given an image of height $H$ and width $W$,
the ResNet50 backbone model extracts a lower-resolution feature maps of dimension $2048\times H/32 \times W/32$. We reduce the dimension of the feature maps to \textit{d} using $1\times1$ convolution and
vectorize the features into \textit{d}$\times HW$ feature vectors. Transformer architecture is permutation-invariant and while processing the feature vectors, the spatial information is lost. 
To tackle this, similar to Transformer architectures for NLP problems, we use fixed positional encoding.
\subsubsection{Transformer encoder}
The supplemented feature vector with the fixed sine positional encoding~\citep{vaswani2017attention} is provided as input to each layer of the encoder. Each encoder layer consists
of multi-headed self-attention with 256-dimensional \textit{query, key, and value} vectors and a feed-forward network (FFN). The self-attention mechanism equipped 
with positional encoding enables learning the spatial relationship between pixels. Unlike CNNs which model the spatial relationship between pixels
in a small fixed neighborhood defined by the kernel size, the self-attention mechanism enables learning spatial relationships between pixels over the entire image.

\subsubsection{Transformer decoder}
On the decoder part, from the encoder output embedding and $N$ positional embedding inputs, we generate $N$ decoder output embeddings using standard multi-head attention mechanism. $N$ is the cardinality of the set we predict. Unlike the fixed sine positional encoding used in the encoder, we use learned positional encoding in the decoder. We call this encoding \textit{object queries}. From the $N$ decoder output embeddings, we use feed-forward prediction heads
to generate set of $N$ output tuples. Note that each tuple in the set is generated from a decoder output embedding independently---lending itself for efficient parallel processing.

\subsubsection{Prediction heads}
For each decoder output (object query), we use four feed forward prediction heads to predict the class probability, bounding box modeled as the center and scale, translation and orientation components of 6D pose independently. Prediction heads are straightforward three-layer MLPs with 256 neurons in each hidden layer.

\section{Experiments}

\subsection{Dataset}
The YCB-Video (YCB-V) dataset~\cite{xiang2017posecnn} is a benchmark dataset for the 6D pose estimation task. The dataset consists of 92 video sequences of random subset of objects from a total of 21 objects arranged in random configurations. 
In total, the dataset consists of 133,936 images in $640 \times 480$ resolution with segmentation masks, depths, bounding boxes, and 6D object pose annotations.
Twelve video sequences are held out for the test set with 20,738 images, and the rest images are used for training.
Additionally, PoseCNN~\cite{xiang2017posecnn} provides 80K synthetic images for training.
For the validation set, we adopt the BOP test set of YCB-V~\cite{hodavn2020bop}, a subset of 75 images from each of the 12 test scenes totaling 900 images.
For the final evaluation, we follow the same approach as~\cite{xiang2017posecnn} and report the results on the subset of 2,949 key frames from 12 test scenes.

\subsection{Metrics}
For the model evaluation, the average distance (ADD) metric is employed from~\cite{hinterstoisser2013mbt}. Given the predicted $\hat{R}$ and $\hat{t}$ and their corresponding ground-truths, ADD calculates the mean pairwise distance between transformed 3D model points ($\mathcal{M}$). If the ADD is below 0.1\,m we consider the pose prediction to be correct.

\begin{equation}
	\text{ADD} = \frac{1}{|\mathcal{M}|} \sum_{x \in \mathcal{M}}\|(Rx+t)-(\hat{R} x+\hat{t})\|
\end{equation}

For symmetric objects, instead of using ADD metric, the average closest pairwise distance (ADD-S) metric is computed as follows:

\begin{equation}
	\text{ADD-S} = \frac{1}{|\mathcal{M}|} \sum_{x_{1} \in \mathcal{M}} \min _{x_{2} \in \mathcal{M}}\|(R x_{1}+t)-(\hat{R} x_{2}+\hat{t})\|
\end{equation}

Following \citep{xiang2017posecnn},  
we aggregate all results and measure the area under the
accuracy-threshold curve (AUC) for distance thresholds of maximum 0.1\,m.

\subsection{Training}\label{sec:train}

The DETR architecture suffers from the drawback of having a slow convergence~\cite{zhu2020deformable}. 
To tackle this issue, we initialize the model using the provided pretrained weights on the COCO dataset~\citep{lin2014microsoft} 
and then train the complete T6D-Direct model on the YCB-V dataset. After initializing our model with the pretrained weights, there are two possible strategies while training for the pose estimation task.
In the first approach, we train the complete model for both object detection and pose estimation tasks simultaneously; 
therefore, the total loss function is the Hungarian loss brought in \cref{hungarian_loss}. 
In the second approach, we employ a multi-stage scheme to train only the pose prediction heads and freeze the rest of the network. Investigation on these methods are conducted in~\cref{sec:ablation}. 

\begin{figure}[t]
        \centering
        \newlength{\imgw}
        \setlength{\imgw}{3.0cm}
        \setlength{\tabcolsep}{0.009cm}
        \begin{tabular}{p{1.em}ccc}
                \raisebox{2.75\normalbaselineskip}[0pt][0pt] {\rotatebox[origin=c]{90}{\scriptsize Scene}} &
         \includegraphics[width=\imgw]{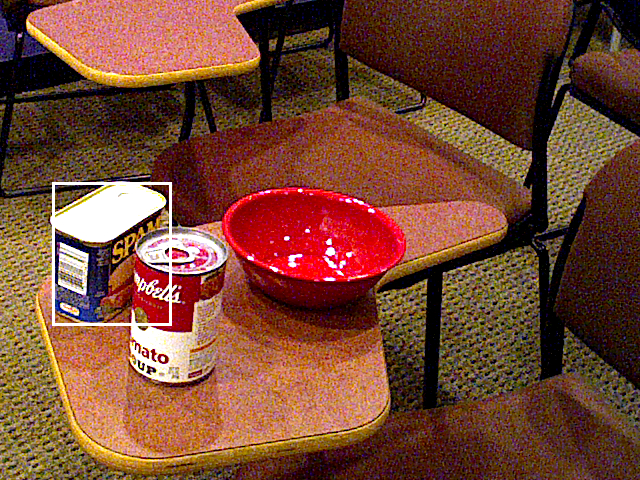} &
         \includegraphics[width=\imgw]{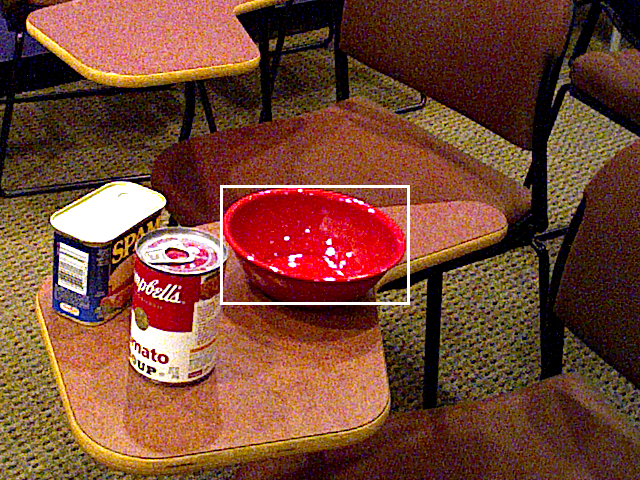} &
         \includegraphics[width=\imgw]{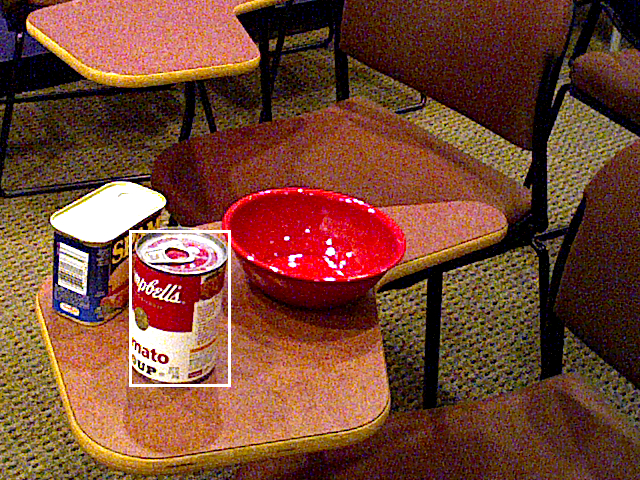}  \\
         \raisebox{2.75\normalbaselineskip}[0pt][0pt] {\rotatebox[origin=c]{90}{\scriptsize Full model}} &
         \includegraphics[width=\imgw]{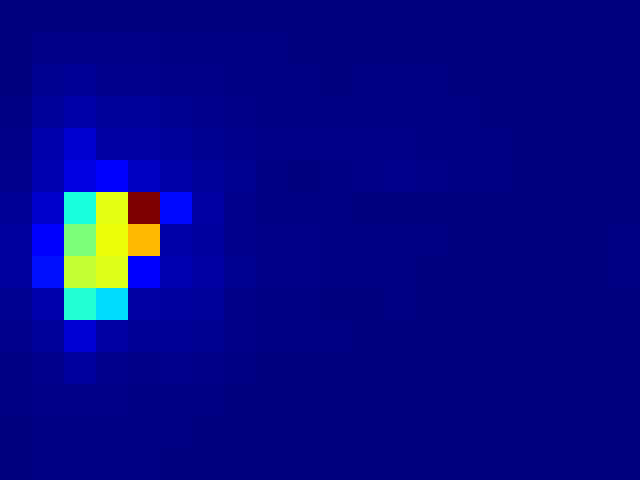} &
         \includegraphics[width=\imgw]{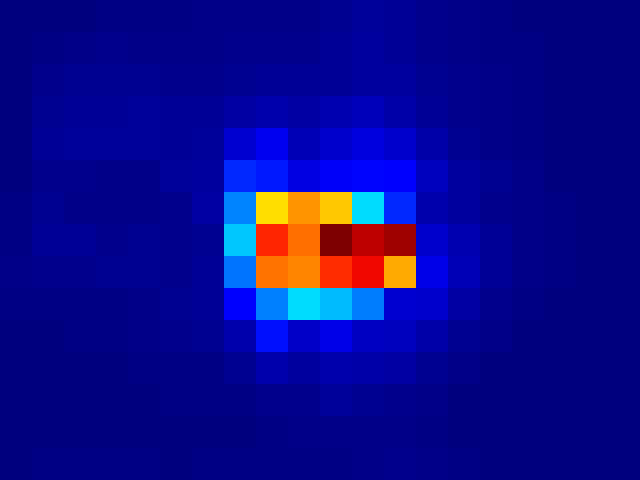} &
         \includegraphics[width=\imgw]{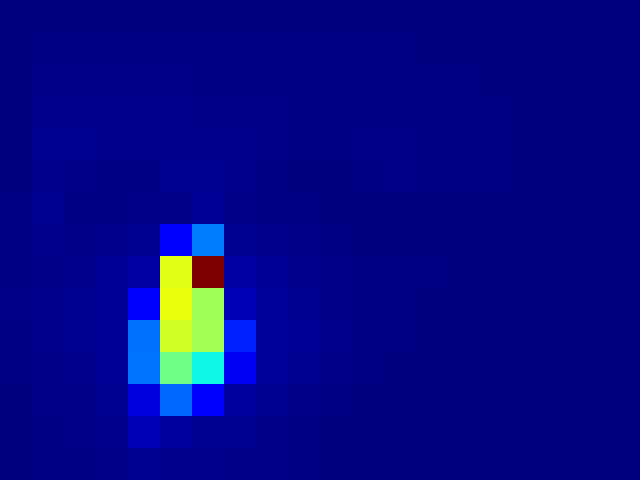} 
          \\
         \raisebox{2.5\normalbaselineskip}[0pt][0pt] {\rotatebox[origin=c]{90}{\scriptsize Frozen}} &
         \includegraphics[width=\imgw]{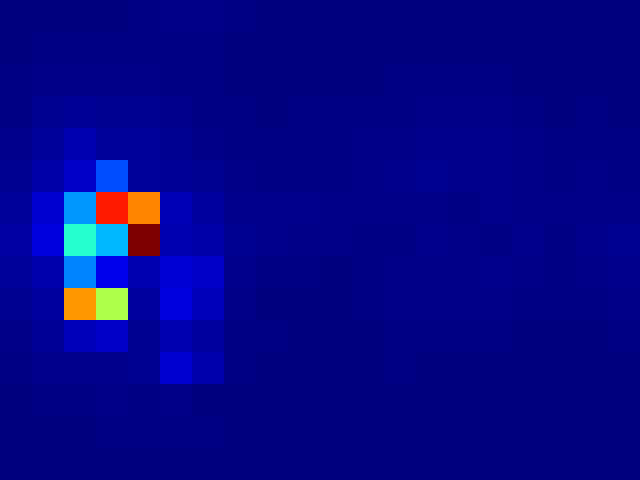} &
         \includegraphics[width=\imgw]{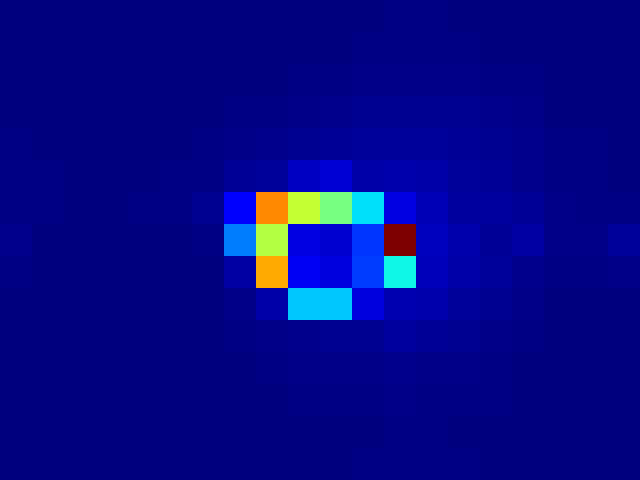} &
         \includegraphics[width=\imgw]{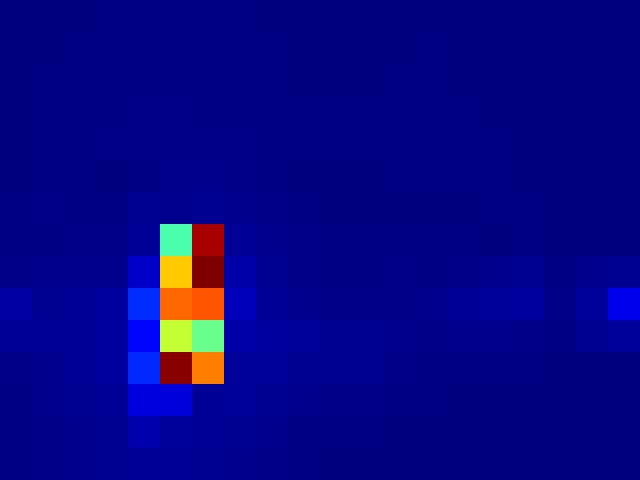} 
         \\
        \end{tabular}
        \tikz{
                \pgfplotscolorbardrawstandalone[ 
                        colormap/jet,
                        point meta min=0,
                        point meta max=1,
                        colorbar horizontal, 
                        colorbar style={
                            height=0.15cm,
                            width=7cm,
                            font=\scriptsize
                        }]
                }
        \caption{Object predictions of a given image (first row) and decoder attention maps for the object queries (second and third rows). 
        Training the complete model for both object detection and pose estimation tasks (second row). Training the model first on the object detection task, and then training the frozen model on the pose estimation task (third row).
        Attention maps are visualized using the jet color map (shown above for reference).}
        \label{fig:multiheadattn}
\end{figure}

To further understand the behavior of the mentioned approaches, we visualize the decoder attention
maps for the object queries corresponding to the predictions. In \cref{fig:multiheadattn}, the top row consists of the object predictions.
The middle and bottom rows consist of the attention maps from the complete and partial trained models, respectively, corresponding to the object predictions in the top row. The partial trained model has higher activations along the object boundaries. 
These activations are the result of training the partial model only on the
object detection task. When freezing the transformer model and training only the prediction heads, the prediction heads have to rely on the features
already learned, whereas the complete trained model has denser activations compared to the partial trained model and the activations are spread
over the whole object and not just the object boundaries. Thus, training the complete model helps learn features more suitable for pose
estimation than the features learned for object detection.

\subsubsection{Hyperparameters}
$\alpha$ and $\beta$ hyperparameters in computing $\mathcal{L}_{box}$ (\cref{eqn:box_loss}) are set to 2 and 5, respectively.
The $\lambda_{pose}$ hyperparameter in computing $\mathcal{L}_{Hungarian}$ (\cref{hungarian_loss}) is set to 0.05, and the cardinality of the predicted set $N$ is set to 20.
The model takes the image of the size 640 $\times$ 480 as input and is trained using the AdamW optimizer~\cite{adamw}
with an initial learning rate of $10^{-4}$ and for 78K iterations. 
The learning rate is decayed to $10^{-5}$ after 70K iterations, and the batch size is 32. Moreover, gradient clipping with a maximal gradient norm of 0.1 is applied.
In addition to YCB-V dataset images, we use the synthetic dataset provided by PoseCNN for training our model.

\subsection{Results}

In \cref{tab:experiments:ycbv}, we present the per object quantitative results of T6D-Direct on the YCB-V dataset.
We compare our results with PoseCNN~\citep{xiang2017posecnn},  PVNet~\citep{peng2019pvnet} and DeepIM~\citep{li2018deepim}. In terms of the approach, T6D-Direct
is comparable to PoseCNN; both are direct regression methods, whereas PVNet is an indirect method, and DeepIM is a refinement-based approach.
In terms of both the AUC of ADDS and AUC of ADD(-S) metrics, T6D-Direct outperforms PoseCNN and outperforms the AUC of ADD(-S) results of PVNet. For a fair comparison, we follow the same object symmetry definition and evaluation procedure described by the YCB-Video dataset~\citep{xiang2017posecnn}.

\begin{figure}[htp]
        \centering
          \resizebox{.9\linewidth}{!}{\input{esa.tikz}}
          \tikz{
                \pgfplotscolorbardrawstandalone[ 
                        colormap/jet,
                        point meta min=0,
                        point meta max=1,
                        colorbar horizontal, 
                        colorbar style={
                            height=0.15cm,
                            width=7cm,
                            font=\scriptsize
                        }]
            }
          \caption{Encoder self-attention. We visualize the self-attention maps for three pixels belonging to three objects
          in the image. All three pixels lie on the same horizontal line but attend to different parts of the image. Attention maps are visualized using the jet color map (shown above for reference).}
          \label{fig:enc_sa}
\end{figure}
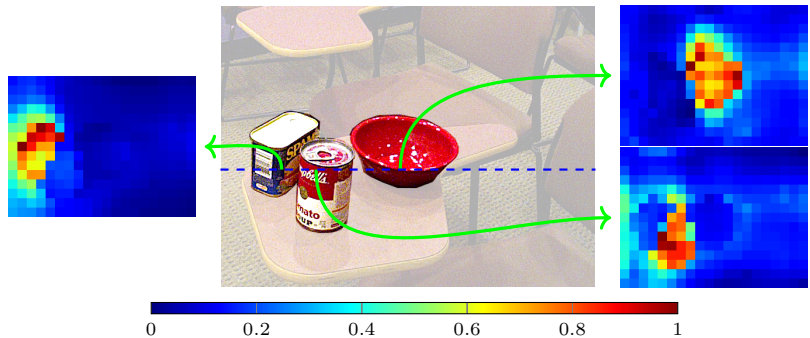

\begin{figure}[h]
	\centering
	\newlength{\imgres}
	\setlength{\imgres}{0.16\textwidth}
	\setlength{\tabcolsep}{0.01cm}
	\begin{tabular}{p{.5em}cccccc}
	 \raisebox{2.75\normalbaselineskip}[0pt][0pt] {\rotatebox[origin=c]{90}{\qquad \tiny Ours \qquad \quad PoseCNN}} &
	 \includegraphics[width=\imgres]{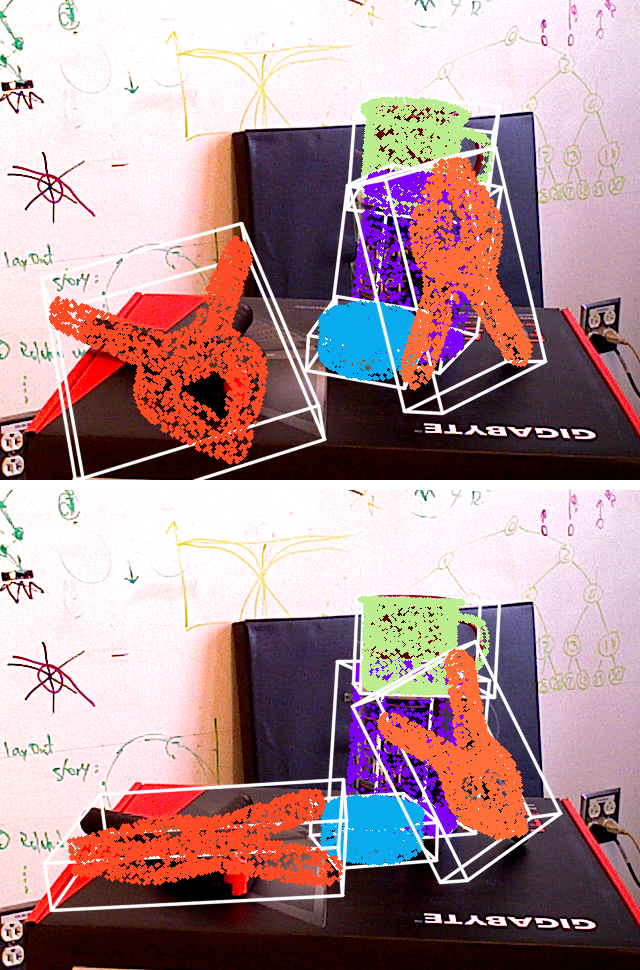} &
	 \includegraphics[width=\imgres]{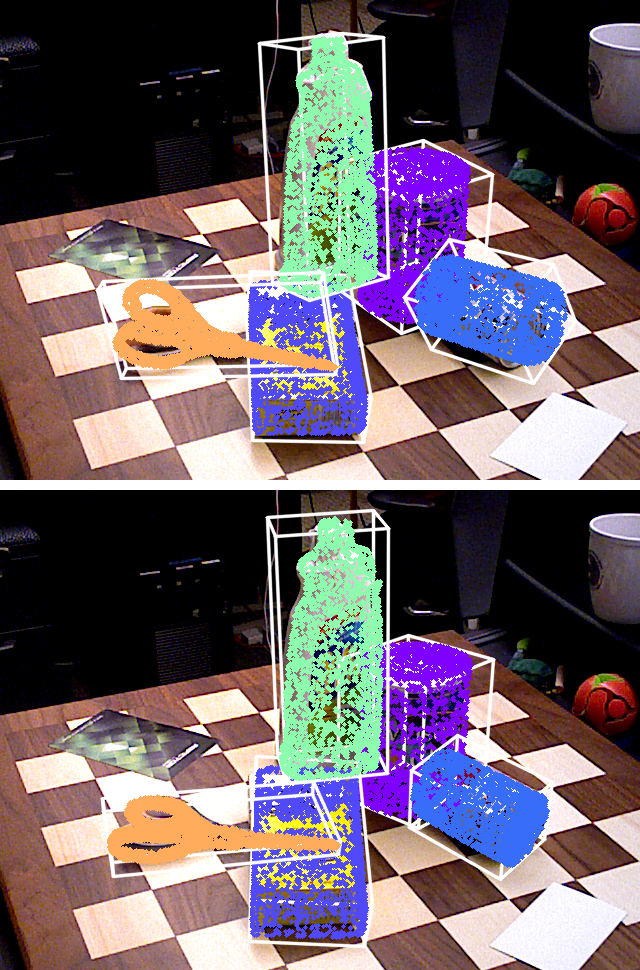} &
	 \includegraphics[width=\imgres]{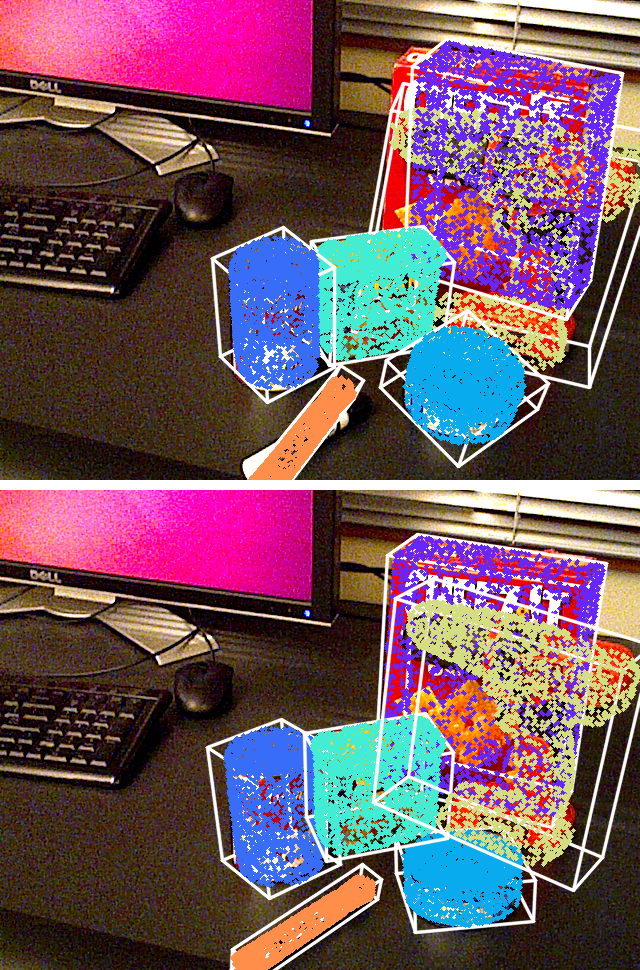} &
	 \includegraphics[width=\imgres]{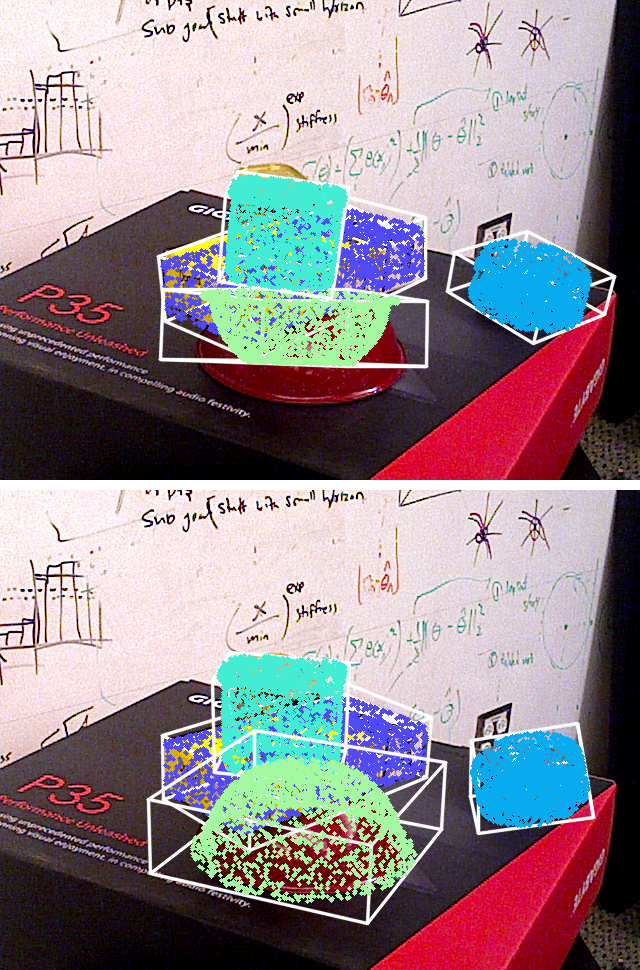} &
	 \includegraphics[width=\imgres]{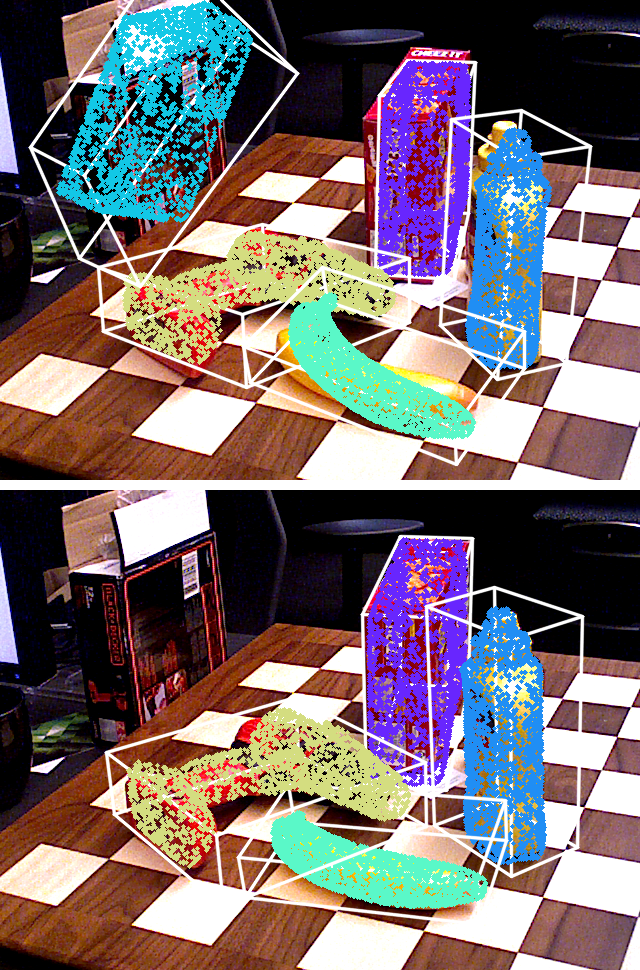} &
	 \includegraphics[width=\imgres]{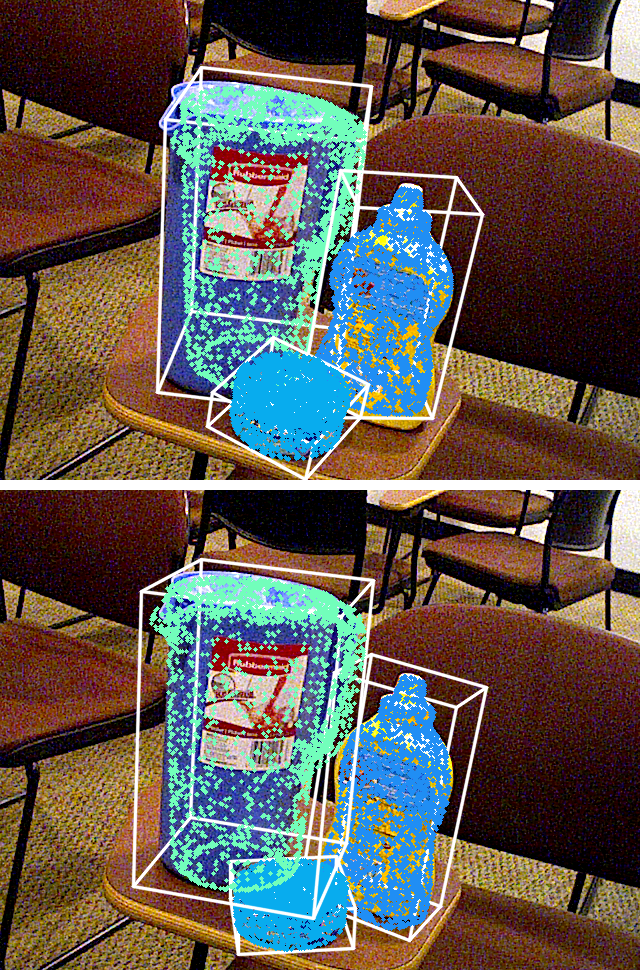}
	\end{tabular}
	\caption{Qualitative examples from the YCB-V Dataset.
	Top row: PoseCNN~\citep{xiang2017posecnn}. Bottom row: our predictions.}
	\label{fig:result}
\end{figure}

Some qualitative results are shown in \cref{fig:result}. To demonstrate the ability of the Transformer architecture to model dependencies between pixels over the whole image instead of a just small local neighborhood, in \cref{fig:enc_sa}, we visualize the self-attention maps for three pixels belonging to three objects in the image. All three pixels lie on the same horizontal line but attend to different parts of the image.

\begin{table}[htp]
  \centering
  \footnotesize
  \caption{Pose prediction results on the YCB-V Dataset. The symmetric objects are denoted by *.}
  \begin{tabular}{lccc|cccc}
    \toprule
    Metric & \multicolumn{3}{c}{AUC of ADD-S} & \multicolumn{4}{c}{AUC of ADD(-S)}  \\
%    \cmidrule(r){2-4}\cmidrule(r){5-8}
%    Refinement & & & \checkmark & & & & \checkmark \\
    \midrule
    Object & \scriptsize PoseCNN & \scriptsize T6D-Direct & \scriptsize DeepIM & \scriptsize PoseCNN & \scriptsize PVNet & \scriptsize T6D-Direct & \scriptsize DeepIM \\
    \bottomrule
    master\_chef\_can & 84.0 & 91.9 & 93.1 & 50.9 & 81.6 & 61.5 & 71.2 \\ 
    cracker\_box & 76.9 & 86.6 & 91.0 & 51.7 & 80.5 & 76.3 & 83.6 \\ 
    sugar\_box & 84.3 & 90.3 & 96.2 & 68.6 & 84.9 & 81.8 & 94.1 \\ 
    tomato\_soup\_can & 80.9 & 88.9 & 92.4 & 66.0 & 78.2 & 72.0 & 86.1 \\
    mustard\_bottle & 90.2 & 94.7 & 95.1 & 79.9 & 88.3 & 85.7 & 91.5 \\
    tuna\_fish\_can & 87.9 & 92.2 & 96.1 & 70.4 & 62.2 & 59.0 & 87.7 \\
    pudding\_box & 79.0 & 85.1 & 90.7 & 62.9 & 85.2 & 72.7 & 82.7 \\
    gelatin\_box & 87.1 & 86.9 & 94.3 & 75.2 & 88.7 & 74.4 & 91.9 \\
    potted\_meat\_can & 78.5 & 83.5 & 86.4 & 59.6 & 65.1 & 67.8 & 76.2 \\
    banana & 85.9 & 93.8 & 72.3 & 91.3 & 51.8 & 87.4 & 81.2 \\
    pitcher\_base & 76.8 & 92.3 & 94.6 & 52.5 & 91.2 & 84.5 & 90.1 \\
    bleach\_cleanser & 71.9 & 83.0 & 90.3 & 50.5 & 74.8 & 65.0 & 81.2 \\
    bowl* & 69.7 & 91.6 & 81.4 & 69.7 & 89.0 & 91.6 & 81.4 \\
    mug & 78.0 & 89.8 & 91.3 & 57.7 & 81.5 & 72.1 & 81.4 \\
    power\_drill & 72.8 & 88.8 & 92.3 & 55.1 & 83.4 & 77.7 & 85.5 \\
    wood\_block* & 65.8 & 90.7 & 81.9 & 65.8 & 71.5 & 90.7 & 81.9 \\
    scissors & 56.2 & 83.0 & 75.4 & 35.8 & 54.8 & 59.7 & 60.9 \\
    large\_marker & 71.4 & 74.9 & 86.2 & 58.0 & 35.8 & 63.9 & 75.6 \\
    large\_clamp* & 49.9 & 78.3 & 74.3 & 49.9 & 66.3 & 78.3 & 74.3 \\
    extra\_large\_clamp* & 47.0 & 54.7 & 73.2 & 47.0 & 53.9 & 54.7 & 73.3 \\
    foam\_brick* & 87.8 & 89.9 & 81.9 & 87.8 & 80.6 & 89.9 & 81.9 \\
    \midrule
    MEAN & 75.9 & 86.2 & 88.1 & 61.3 & 73.4 & 74.6 & 81.9 \\
    \bottomrule
  \end{tabular}
  \label{tab:experiments:ycbv}
\end{table}

\subsection{Inference time analysis}

\begin{table}[h]
    %  \hspace{0.01\textwidth}
      \centering
      \footnotesize
      \setlength{\aboverulesep}{0pt}
      \setlength{\belowrulesep}{0pt}
        \caption{Comparison with state-of-the-art methods on YCB-V. In terms of the ADD(-S) 0.1d metric, we achieve the state-of-the-art result. $^{\dagger}$ indicates that the method is refinement-based. Inference time is the average time taken for processing all objects in an image.}
\begin{tabular}{l|c|c|c|c}
    \toprule
    Method & ADD(-S) & \thead{AUC of \\ ADD-S} & \thead{AUC of \\ ADD(-S)} & \thead{Inference Time \\{} [\si{\s}] }\\
    \bottomrule
    CosyPose$^{\dagger}$~\citep{labbe2020} & - & \textbf{89.8} & \textbf{84.5} & 0.395 \\
    PoseCNN~\citep{xiang2017posecnn} & 21.3 & 75.9 & 61.3 & *$>$0.250 \\
    SegDriven~\citep{hu2019segmentation} & 39.0 & - & - & *0.050 \\
    Single-Stage~\citep{Hu2020CVPR} & \textbf{53.9} & - & - & *0.022 \\
    GDR-Net~\citep{Wang_2021_GDRN} & 49.1 & 89.1 & 80.2 & 0.065 \\
    T6D-Direct (Ours) & 48.7 & 86.2 & 74.6 & \textbf{0.017} \\
    \bottomrule
  \end{tabular}
\end{table}

Since the prediction heads generate $N$ predictions in parallel, the inference of our model is not dependent on the number of objects in an image. 
However, having a smaller cardinality of the prediction set requires estimating fewer object queries and facilitates faster inference time. Thus, we set N to 20. On an NVIDIA 3090 GPU and Intel 3.70\,GHz CPU, our model runs at 58\,fps which makes our model ideal for real-time applications.

\section{Ablation study}\label{sec:ablation}

In this section, we explore the effect of various training strategies, different loss functions, and egocentric \textit{vs.} allocentric rotation representations on the T6D-Direct model performance for the YCB-V dataset.

\subsubsection{Effectiveness of loss functions}
In~\cref{table:ablation}, we examine the performance of our model using the symmetry aware version of Point Matching loss with $\ell_2$ norm~\citep{li2018deepim, xiang2017posecnn} which, in contrast to the disentangled loss presented in~\cref{sec:poseloss}, couples the rotation and translation components. This loss function results in the best AUC of ADD(-S) metric.
Moreover, since the symmetry aware SLoss component of the Point Matching loss is computationally expensive, we experimented with training our model using only the PLoss component. Interestingly, the ADD(-S) result of the model trained using only the PLoss component (row 5) is only slightly worse than the model trained using the both components (row 1).

\begin{table}[h]
  \centering
  \footnotesize
  \caption{Ablation study on YCB-V. We provide results of our method with different loss functions and training schemes.}
  \begin{tabular}{c|l|c|c}
    \toprule
    Row & Method & ADD(-S) & \thead{AUC of \\ ADD(-S)} \\
    \bottomrule
    1 & T6D-Direct with Point Matching loss & 47.0 & \textbf{75.6} \\
    2 &  1 + multi-stage training & 20.5 & 59.1 \\
    3 &  1 + pose matching cost component & 42.8 & 71.7 \\
    4 & 1 + allocentric R$_{6d}$ & 42.9 & 74.4 \\
    5 & T6D-Direct with PLoss & 45.8 & 74.4 \\
    6 & T6D-Direct & \textbf{48.7} & 74.6 \\
    \bottomrule
  \end{tabular}
  \label{table:ablation}
\end{table}

\subsubsection{Effectiveness of training strategies}
%The loss function we use is based on the PoseLoss and
%ShapeMatch-Loss from PoseCNN[43] but instead of considering only the rotation, our approach takes also the translation into %account. 

As discussed in~\cref{sec:train}, there are two training schemes: single-stage and multi-stage. In the multi-stage scheme, we train the Transformer model for object detection and only train the FFNs for pose estimation, whereas in the single-stage scheme, we train the complete model in one stage. In our experiments, as shown in~\cref{table:ablation}, multi-stage training (row 2) yielded inferior results, although both schemes were pretrained on the COCO dataset. This demonstrates that the Transformer model is learning the features specific to the 6D object pose estimation task on YCB-V, and COCO fine-tuning mainly helps in faster convergence during training and not in more accurate pose estimations. We thus believe that most large-scale image datasets can serve as pretraining data source. We also provide the results of including the pose component in the bipartite matching cost mentioned in~\cref{match_pose}. Including the pose component (row 3) does not provide any considerable advantage; thus, we include only the class probability and bounding box components in the bipartite matching cost in all further experiments. Further, egocentric rotation representation (row 1) performed slightly better than allocentric representation (row 4). We hypothesize that supplementing RGB images with positional encoding allows the Transformer model to learn spatial features efficiently. Therefore, the allocentric representation does not have any advantage over the egocentric representation.

\section{Conclusion}
We introduced T6D-Direct, a transformer-based architecture for multi-object 6D pose estimation.
Equipped with multi-head attention mechanism, our model obtains competitive results in the task of direct 6D pose estimation without any dense features.
Unlike the standard multi-staged methods, our formulation of multi-object 6D pose estimation as a set prediction problem allows
estimating the 6D pose of all the objects in a given image in one forward pass.
Furthermore, our model is real-time capable. In the future, we plan to explore the possibilities of incorporating
dense estimation features into our architecture and improve the performance further.

\section*{Acknowledgment}

This research has been supported by the Competence Center for Machine Learning Rhine Ruhr
(ML2R), which is funded by the Federal Ministry of Education and Research of Germany (grant no. 01—S18038A).

\bibliography{references}

\end{document}

%% file: pipeline.tikz
\begin{tikzpicture}

\node (canvas) at (0, 0) {\includegraphics[width=0.99\linewidth]{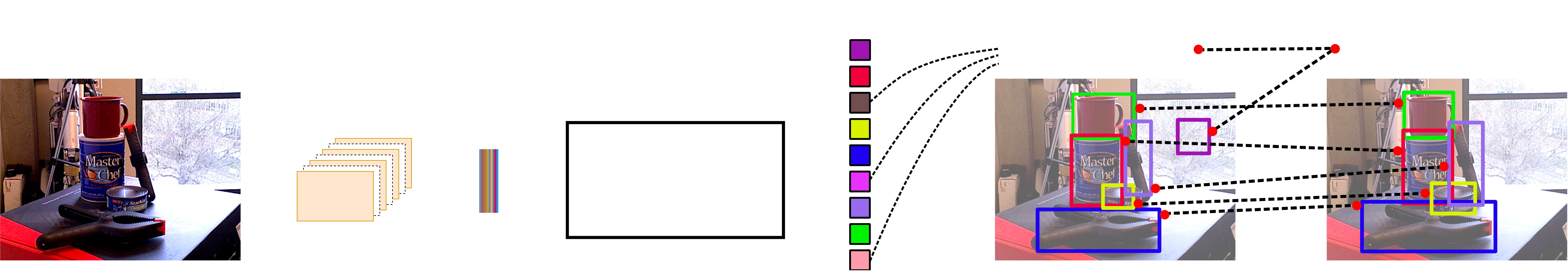}};

\node [align=center] at (-0.8, -0.1){ \scriptsize Transformer };
\node [align=center] at (-0.8, -0.40){ \scriptsize  encoder-};
\node [align=center] at (-0.8, -0.60){ \scriptsize  decoder};

\node [align=center] at (2.4, 0.65){ \tiny  no object (\O)};
\node [align=center] at (5.1, 0.65){ \tiny  no object (\O)};

\draw[->,  thick] (-4.19,-0.35) -- (-3.75,-0.35);
\draw[->,  thick] (-2.87,-0.35) -- (-2.35,-0.35);
\draw[->,  thick] (-2.2,-0.35) -- (-1.68,-0.35);
\draw[->,  thick] (0,-0.35) -- (.5,-0.35);

\node[text=black, thick]  at (-5.1, 0.6){\scriptsize $H \mathsf{x} W$};
\node [align=center] at (-3.2, -1.4)[align=center, font=\tiny\linespread{1.2}\selectfont]{\scriptsize CNN \scriptsize features};
\node [align=center,text=black, thick,] at (-3.6, .52){\scalebox{.7} {{\scriptsize 2048}$\mathsf{x}\frac{H}{32}\mathsf{x}\frac{W}{32}$}};

\node [align=center,text=black, thick] at (-2.33, .52){\scalebox{.7} { {\scriptsize 256}$\mathsf{x}(\frac{H}{32}\mathsf{*}\frac{W}{32})$}};

\node [align=center] at (2.55, -1.2){\scriptsize  Set of predictions};
\node [align=center] at (5.1, -1.2){\scriptsize  Ground truth};

\rotateRPY{20}{0}{0}
    \begin{scope}[RPY]
       \coordinate (O1) at (2.5, 0.1,0);
\draw[red, thick,-] ( $ (O1) + (0,0,0) $) -- ( $ (O1) + (0.2,0,0) $) node[anchor=north east]{};
\draw[green, thick,-] ( $ (O1) + (0,0,0) $)  -- ( $ (O1) +(0,0.2,0) $) node[anchor=north west]{};
\draw[blue, thick,-] ( $ (O1) + (0,0,0) $)  -- ( $ (O1) +(0,0,0.2) $) node[anchor=south]{}; 
    \end{scope}

\rotateRPY{40}{00}{0}
    \begin{scope}[RPY]
       \coordinate (O1) at (2.3, -0.4, 0);
\draw[red, thick,-] ( $ (O1) + (0,0,0) $) -- ( $ (O1) + (0.2,0,0) $) node[anchor=north east]{};
\draw[green, thick,-] ( $ (O1) + (0,0,0) $)  -- ( $ (O1) +(0,0.2,0) $) node[anchor=north west]{};
\draw[blue, thick,-] ( $ (O1) + (0,0,0) $)  -- ( $ (O1) +(0,0,0.2) $) node[anchor=south]{}; 
    \end{scope}

\rotateRPY{-30}{00}{0}
    \begin{scope}[RPY]
       \coordinate (O1) at (2.55, -0.75, 0);
\draw[red, thick,-] ( $ (O1) + (0,0,0) $) -- ( $ (O1) + (0.2,0,0) $) node[anchor=north east]{};
\draw[green, thick,-] ( $ (O1) + (0,0,0) $)  -- ( $ (O1) +(0,0.2,0) $) node[anchor=north west]{};
\draw[blue, thick,-] ( $ (O1) + (0,0,0) $)  -- ( $ (O1) +(0,0,0.2) $) node[anchor=south]{}; 
    \end{scope}

\rotateRPY{-10}{00}{0}
    \begin{scope}[RPY]
       \coordinate (O1) at (2.7, -0.3, 0);
\draw[red, thick,-] ( $ (O1) + (0,0,0) $) -- ( $ (O1) + (0.2,0,0) $) node[anchor=north east]{};
\draw[green, thick,-] ( $ (O1) + (0,0,0) $)  -- ( $ (O1) +(0,0.2,0) $) node[anchor=north west]{};
\draw[blue, thick,-] ( $ (O1) + (0,0,0) $)  -- ( $ (O1) +(0,0,0.2) $) node[anchor=south]{}; 
    \end{scope}

\rotateRPY{50}{00}{0}
    \begin{scope}[RPY]
       \coordinate (O1) at (2.5,-0.5, 0);
\draw[red, thick,-] ( $ (O1) + (0,0,0) $) -- ( $ (O1) + (0.2,0,0) $) node[anchor=north east]{};
\draw[green, thick,-] ( $ (O1) + (0,0,0) $)  -- ( $ (O1) +(0,0.2,0) $) node[anchor=north west]{};
\draw[blue, thick,-] ( $ (O1) + (0,0,0) $)  -- ( $ (O1) +(0,0,0.2) $) node[anchor=south]{}; 
    \end{scope}

\rotateRPY{-50}{00}{0}
    \begin{scope}[RPY]
       \coordinate (O1) at (3.1,-0.05, 0);
\draw[red, thick,-] ( $ (O1) + (0,0,0) $) -- ( $ (O1) + (0.2,0,0) $) node[anchor=north east]{};
\draw[green, thick,-] ( $ (O1) + (0,0,0) $)  -- ( $ (O1) +(0,0.2,0) $) node[anchor=north west]{};
\draw[blue, thick,-] ( $ (O1) + (0,0,0) $)  -- ( $ (O1) +(0,0,0.2) $) node[anchor=south]{}; 
    \end{scope}

\rotateRPY{0}{0}{0}
    \begin{scope}[RPY]
       \coordinate (O1) at (5,0.1,0);
\draw[red, thick,-] ( $ (O1) + (0,0,0) $) -- ( $ (O1) + (0.2,0,0) $) node[anchor=north east]{};
\draw[green, thick,-] ( $ (O1) + (0,0,0) $)  -- ( $ (O1) +(0,0.2,0) $) node[anchor=north west]{};
\draw[blue, thick,-] ( $ (O1) + (0,0,0) $)  -- ( $ (O1) +(0,0,0.2) $) node[anchor=south]{}; 
    \end{scope}

\rotateRPY{10}{0}{0}
    \begin{scope}[RPY]
       \coordinate (O1) at (5, -0.2,0);
\draw[green, thick,-] ( $ (O1) + (0,0,0) $) -- ( $ (O1) + (0.2,0,0) $) node[anchor=north east]{};
\draw[red, thick,-] ( $ (O1) + (0,0,0) $)  -- ( $ (O1) +(0,0.2,0) $) node[anchor=north west]{};
\draw[blue, thick,-] ( $ (O1) + (0,0,0) $)  -- ( $ (O1) +(0,0,0.2) $) node[anchor=south]{}; 
    \end{scope}

\rotateRPY{0}{0}{0}
    \begin{scope}[RPY]
       \coordinate (O1) at (5.3,-0.3,0);
\draw[green, thick,-] ( $ (O1) + (0,0,0) $) -- ( $ (O1) + (0.2,0,0) $) node[anchor=north east]{};
\draw[blue, thick,-] ( $ (O1) + (0,0,0) $)  -- ( $ (O1) +(0,0.2,0) $) node[anchor=north west]{};
\draw[red, thick,-] ( $ (O1) + (0,0,0) $)  -- ( $ (O1) +(0,0,0.2) $) node[anchor=south]{}; 
    \end{scope}

\rotateRPY{0}{0}{0}
    \begin{scope}[RPY]
       \coordinate (O1) at (5.,-0.8,0);
\draw[green, thick,-] ( $ (O1) + (0,0,0) $) -- ( $ (O1) + (0.2,0,0) $) node[anchor=north east]{};
\draw[blue, thick,-] ( $ (O1) + (0,0,0) $)  -- ( $ (O1) +(0,0.2,0) $) node[anchor=north west]{};
\draw[red, thick,-] ( $ (O1) + (0,0,0) $)  -- ( $ (O1) +(0,0,0.2) $) node[anchor=south]{}; 
    \end{scope}

\rotateRPY{0}{0}{0}
    \begin{scope}[RPY]
       \coordinate (O1) at (5.1,-0.5,0);
\draw[green, thick,-] ( $ (O1) + (0,0,0) $) -- ( $ (O1) + (0.2,0,0) $) node[anchor=north east]{};
\draw[blue, thick,-] ( $ (O1) + (0,0,0) $)  -- ( $ (O1) +(0,0.2,0) $) node[anchor=north west]{};
\draw[red, thick,-] ( $ (O1) + (0,0,0) $)  -- ( $ (O1) +(0,0,0.2) $) node[anchor=south]{}; 
    \end{scope}

\end{tikzpicture} 

%% file: architecture.tikz
\begin{tikzpicture}

\pgfmathsetmacro{\EITH}{1.15}
\node (inimage) at (0, 0.7) {\includegraphics[width=0.1\linewidth]{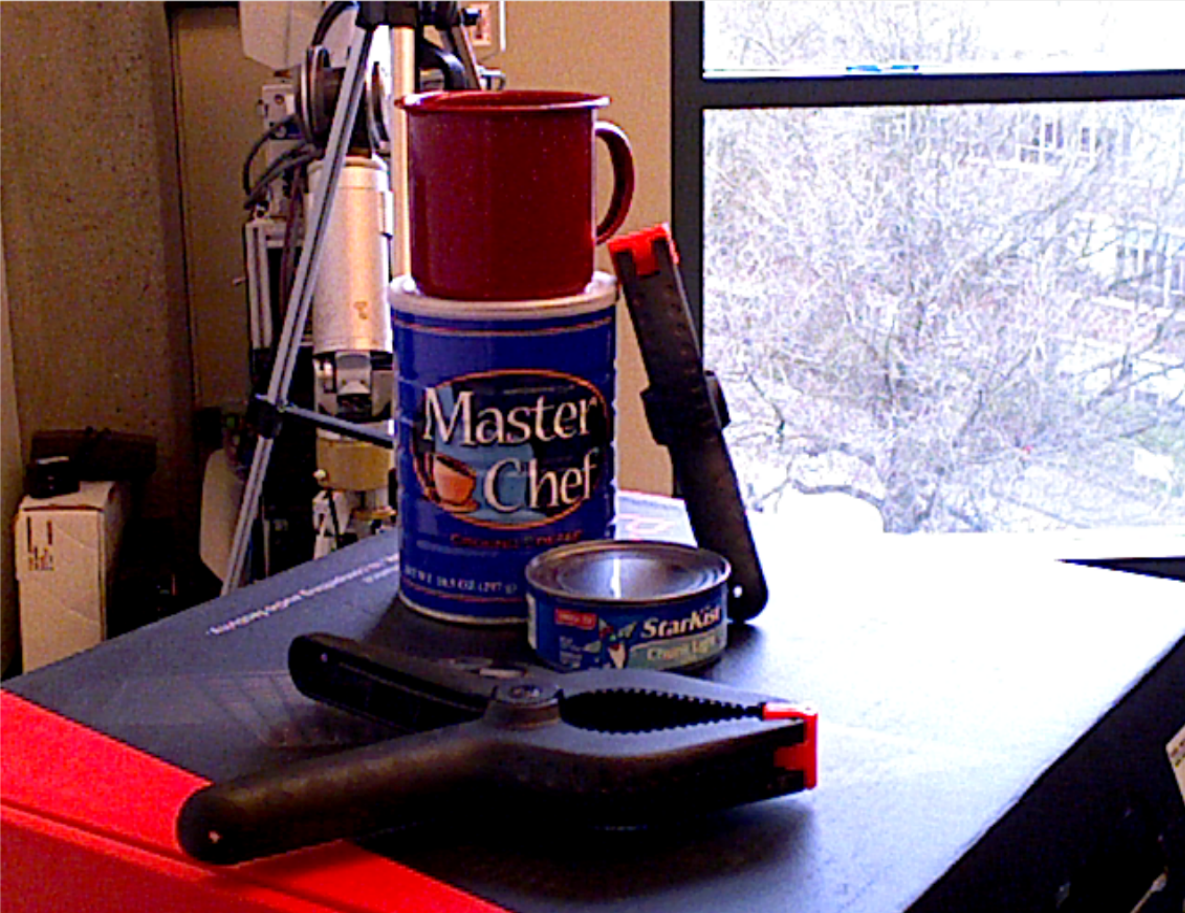}};
\node(cnn) at (1.25, 0.7) [trapezium, trapezium angle=55, rotate=-90, minimum width=10mm, draw, thick, ] {\tiny CNN};
\node (feat) at (1.9, 0.7) [draw,thick,minimum width=0.2cm,minimum height=0.2cm, rotate=-90] {\tiny features};
\coordinate (concat) at (2.5, -0.25);
\node[circle,draw=black, thick, fill=white, inner sep=0pt,minimum size=8pt] (concatcircle) at (concat) {};
\node (backbone) at (0.7, 1.6) [] {\small backbone};
\node (posenc) at (1.2, -1.7) [align=center, font=\tiny\linespread{0.7}\selectfont]{\tiny positional \\ \tiny encoding};
\draw (concat)[very thick] node[cross=4pt,rotate=45]{};
\node (poscode) at (1.2, -0.85) {\includegraphics[width=0.1\linewidth]{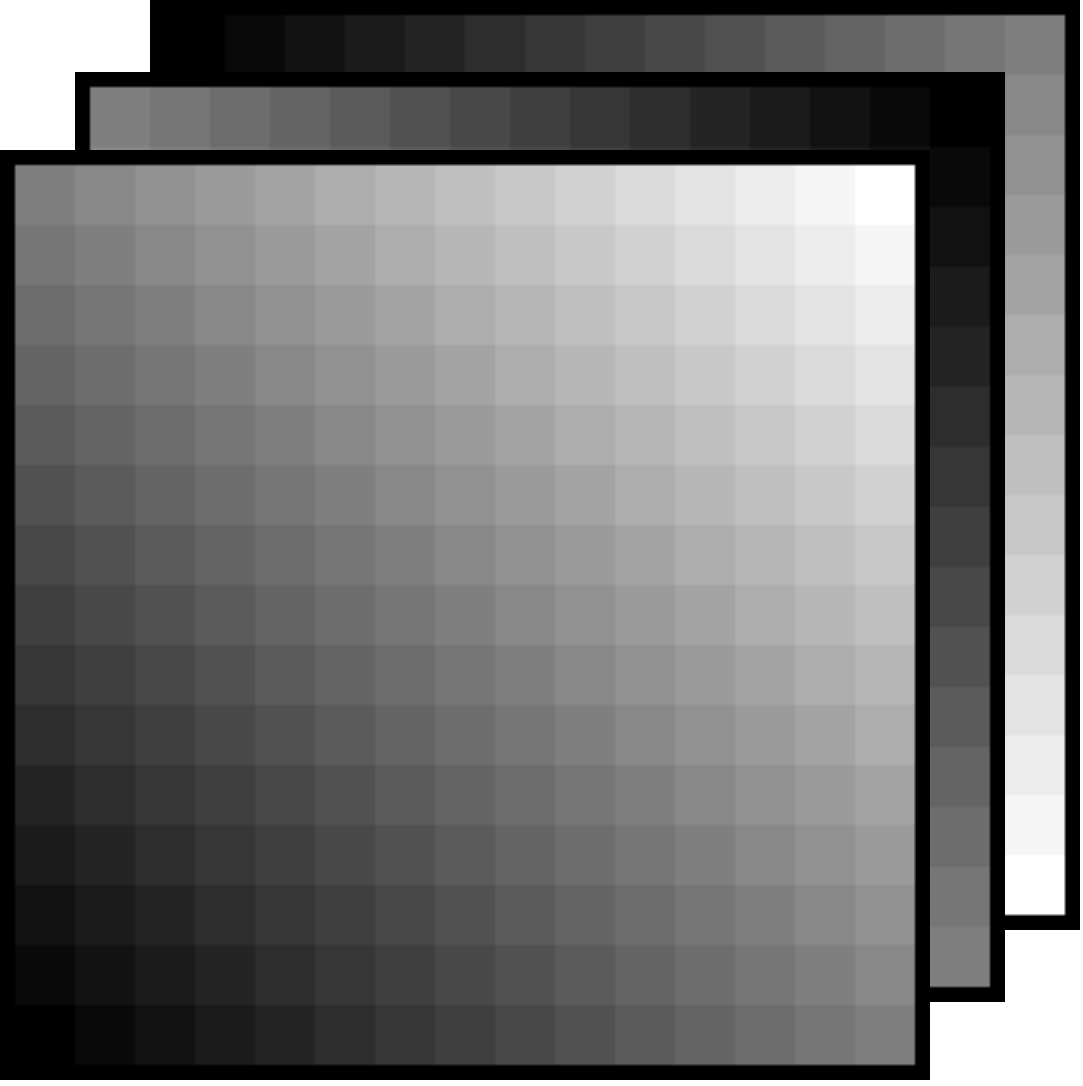}};

\draw [thick, ->](inimage.east) -- (cnn.south);
\draw [thick, ->](cnn.north) -- (feat.south);
\draw [thick, ->](feat.north) -- (concatcircle.135);
\draw [thick, ->](poscode.east) -- (concatcircle.225);

\node(dummy0) at (0, 1.7) {};
\node[draw,  dash pattern={on 2pt off 1pt on 2pt off 1pt},fit=(inimage) (cnn) (feat) (backbone) (dummy0)] {};

\node (ib1) at (3.15, -1.6) [draw, fill=gray,minimum  size=2 pt, scale=0.75] {};
\node (ib2) at (3.4, -1.6) [draw, fill=gray,minimum  size=2 pt, scale=0.75] {};
\node (ib3) at (3.65, -1.6) [draw, fill=gray,minimum  size=2 pt, scale=0.75] {};
\node (ib4) at (3.9, -1.6) [draw, fill=gray,minimum  size=2 pt, scale=0.75] {};
\node (ib5) at (4.15, -1.6) [draw, fill=gray,minimum  size=2 pt, scale=0.75] {};
\node (ib6) at (4.4, -1.6) [draw, fill=gray,minimum  size=2 pt, scale=0.75] {};
\node (ib7) at (4.7, -1.6) [minimum  size=2 pt, scale=0.75] {...};
\node (ib8) at (5., -1.6) [draw, fill=gray,minimum  size=2 pt, scale=0.75]{};

\node (tenc) at (4.1, -0.25) [draw,thick,minimum width=2.5cm,minimum height=1.5cm, align=center] { transformer \\ encoder};

\draw[thick, ->] (ib1.north) -- (ib1.north|-tenc.south);
\draw[thick, ->] (ib2.north) -- (ib2.north|-tenc.south);
\draw[thick, ->] (ib3.north) -- (ib3.north|-tenc.south);
\draw[thick, ->] (ib4.north) -- (ib4.north|-tenc.south);
\draw[thick, ->] (ib5.north) -- (ib5.north|-tenc.south);
\draw[thick, ->] (ib6.north) -- (ib6.north|-tenc.south);
\draw[thick, ->] (ib8.north) -- (ib8.north|-tenc.south);
\node(encfit)[draw, rounded corners=.05cm, fit=(ib1) (ib8) ] {};

\draw [thick, ->] (concatcircle) to[out=-75, in=190] (encfit.west);

\node (it1) at (3.15, \EITH) [draw, fill=gray,minimum  size=2 pt, scale=0.75] {};
\node (it2) at (3.4,  \EITH) [draw, fill=gray,minimum  size=2 pt, scale=0.75] {};
\node (it3) at (3.65, \EITH) [draw, fill=gray,minimum  size=2 pt, scale=0.75] {};
\node (it4) at (3.9, \EITH) [draw, fill=gray,minimum  size=2 pt, scale=0.75] {};
\node (it5) at (4.15, \EITH) [draw, fill=gray,minimum  size=2 pt, scale=0.75] {};
\node (it6) at (4.4, \EITH) [draw, fill=gray,minimum  size=2 pt, scale=0.75] {};
\node (it7) at (4.7, \EITH) [minimum  size=2 pt, scale=0.75] {...};
\node (it8) at (5., \EITH) [draw, fill=gray,minimum  size=2 pt, scale=0.75]{};
\node(encfittop)[draw, rounded corners=.05cm, fit=(it1) (it8) ] {};

\draw[thick, <-] (it1.south) -- (it1.north|-tenc.north);
\draw[thick, <-] (it2.south) -- (it2.north|-tenc.north);
\draw[thick, <-] (it3.south) -- (it3.north|-tenc.north);
\draw[thick, <-] (it4.south) -- (it4.north|-tenc.north);
\draw[thick, <-] (it5.south) -- (it5.north|-tenc.north);
\draw[thick, <-] (it6.south) -- (it6.north|-tenc.north);
\draw[thick, <-] (it8.south) -- (it8.north|-tenc.north);

\node (encoder) at (4.2, 1.6) [] {\small encoder};
\node(dummy1) at (5, -1.7) {};
\node(dummy2) at (5,  1.7) {};
\node[draw,   dash pattern={on 2pt off 1pt on 2pt off 1pt}, fit=(it1) (ib8) (tenc)(dummy2)(dummy1)] {};

\node (ob1) at (6.1, -1.5) [draw, fill=red,minimum  size=2 pt, scale=0.75] {};
\node (ob2) at (6.6, -1.5) [draw, fill=green,minimum  size=2 pt, scale=0.75] {};
\node (ob3) at (7.1, -1.5) [draw, fill=blue,minimum  size=2 pt, scale=0.75] {};
\node (ob4) at (7.6, -1.5) [draw, fill=orange,minimum  size=2 pt, scale=0.75]{};
\node (ob5) at (8.1, -1.5) [draw, fill=violet,minimum  size=2 pt, scale=0.75]{};
\node (ot1) at (6.1, \EITH) [draw, fill=red,minimum  size=2 pt, scale=0.75] {};
\node (ot2) at (6.6, \EITH) [draw, fill=green,minimum  size=2 pt, scale=0.75] {};
\node (ot3) at (7.1, \EITH) [draw, fill=blue,minimum  size=2 pt, scale=0.75] {};
\node (ot4) at (7.6,   \EITH) [draw, fill=orange,minimum  size=2 pt, scale=0.75]{};
\node (ot5) at (8.1,   \EITH) [draw, fill=violet,minimum  size=2 pt, scale=0.75]{};
\node (tdec) at (7.1, -0.25) [draw,thick,minimum width=2.5cm,minimum height=1.5cm, align=center] { transformer \\ decoder};
\draw[thick, <-] (ot1.south) -- (ot1.north|-tdec.north);
\draw[thick, <-] (ot2.south) -- (ot2.north|-tdec.north);
\draw[thick, <-] (ot3.south) -- (ot3.north|-tdec.north);
\draw[thick, <-] (ot4.south) -- (ot4.north|-tdec.north);
\draw[thick, <-] (ot5.south) -- (ot5.north|-tdec.north);
\draw[thick, ->] (ob1.north) -- (ob1.north|-tdec.south);
\draw[thick, ->] (ob2.north) -- (ob2.north|-tdec.south);
\draw[thick, ->] (ob3.north) -- (ob3.north|-tdec.south);
\draw[thick, ->] (ob4.north) -- (ob4.north|-tdec.south);
\draw[thick, ->] (ob5.north) -- (ob5.north|-tdec.south);

\draw [thick, ->] (encfittop.east) to[out=60, in=150] (tdec.west);

\node (encoder) at (7, 1.6) [] {\small decoder};
\node (encoder) at (7.1, -1.8) [] {\small object queries};
\node(dummy3) at (5.9, -1.7) {};
\node(dummy4) at (8.3,  1.7) {};
\node[draw,   dash pattern={on 2pt off 1pt on 2pt off 1pt}, fit=(dummy3)(dummy4)] {};

\node (feat1) at (9.6, 1.15) [draw,thick,minimum width=0.2cm,minimum height=0.2cm] {\tiny FFNs};
\node (feat2) at (9.6, 0.47) [draw,thick,minimum width=0.2cm,minimum height=0.2cm] {\tiny FFNs};
\node (feat3) at (9.6, -0.22) [draw,thick,minimum width=0.2cm,minimum height=0.2cm] {\tiny FFNs};
\node (feat4) at (9.6, -0.9) [draw,thick,minimum width=0.2cm,minimum height=0.2cm] {\tiny FFNs};
\node (feat5) at (9.6, -1.58) [draw,thick,minimum width=0.2cm,minimum height=0.2cm] {\tiny FFNs};

\node (feat6) at (10.9, 1.15) [draw=red,very thick, minimum width=0.2cm,minimum height=0.2cm, align=center, font=\tiny\linespread{0.1}\selectfont,rounded corners=.05cm,inner sep=2pt] {\tiny class \\ \tiny box \\ \tiny pose};
\node (feat7) at (10.9, 0.47) [draw=green,very thick,minimum width=0.2cm,minimum height=0.2cm, align=center,font=\tiny\linespread{0.1}\selectfont,rounded corners=.05cm,inner sep=2pt] {\tiny class \\ \tiny box \\ \tiny pose};
\node (feat8) at (10.9, -0.22) [draw=blue,very thick,minimum width=0.2cm,minimum height=0.2cm, align=center, font=\tiny\linespread{0.1}\selectfont, rounded corners=.05cm,inner sep=2pt] {\tiny class \\ \tiny box \\ \tiny pose};
\node (feat9) at (10.9, -0.9) [draw=orange,very thick,minimum width=0.2cm,minimum height=0.2cm, align=center, font=\tiny\linespread{0.1}\selectfont, rounded corners=.05cm,inner sep=2pt] {\tiny class \\ \tiny box  \\ \tiny pose};
\node (feat10) at (10.9, -1.58) [draw=violet,very thick,minimum width=0.2cm,minimum height=0.2cm, align=center, font=\tiny\linespread{0.1}\selectfont, rounded corners=.025cm,inner sep=2pt] {\tiny class \\ \tiny box  \\ \tiny pose};

\node(dummy5) at (9.25, -1.7) {};
\node(dummy6) at (11.3,  1.7) {};
\node (encoder) at (10.3, 1.6) [] {\small prediction heads};
\node[draw,   dash pattern={on 2pt off 1pt on 2pt off 1pt}, fit=(dummy5)(dummy6)] {};

\draw [thick, ->, red]    (ot1.north) to[out=90, in=150] (feat1.west);
\draw [thick, ->, green]    (ot2.north) to[out=45, in=140] (feat2.west);
\draw [thick, ->, blue]    (ot3.north) to[out=40, in=130] (feat3.west);
\draw [thick, ->, orange]    (ot4.north) to[out=30, in=150] (feat4.west);
\draw [thick, ->, violet]    (ot5.north) to[out=30, in=150] (feat5.west);

\draw [thick, ->, ]     (feat1.east) -- (feat6);
\draw [thick, ->, ]     (feat2.east) -- (feat7);
\draw [thick, ->, ]     (feat4.east) -- (feat9);
\draw [thick, ->, ]     (feat3.east) -- (feat8);
\draw [thick, ->, ]     (feat5.east) -- (feat10);

\end{tikzpicture}

%% file: esa.tikz
\begin{tikzpicture}

\node (scene) at (0, 0) {\includegraphics[width=0.4\linewidth]{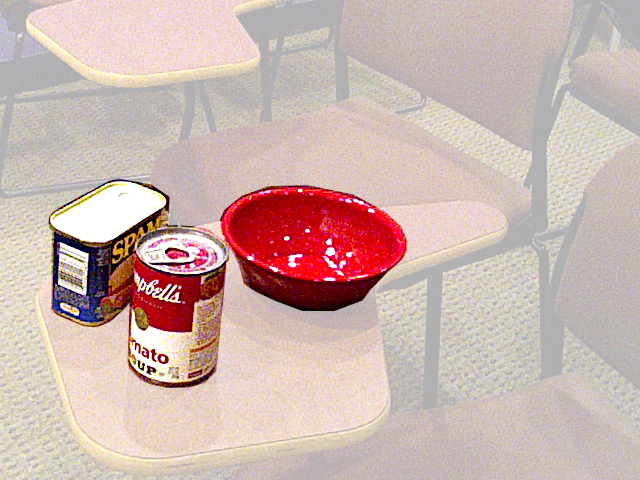}};

\node (sa1) at (-4, 0) {\includegraphics[width=0.2\linewidth]{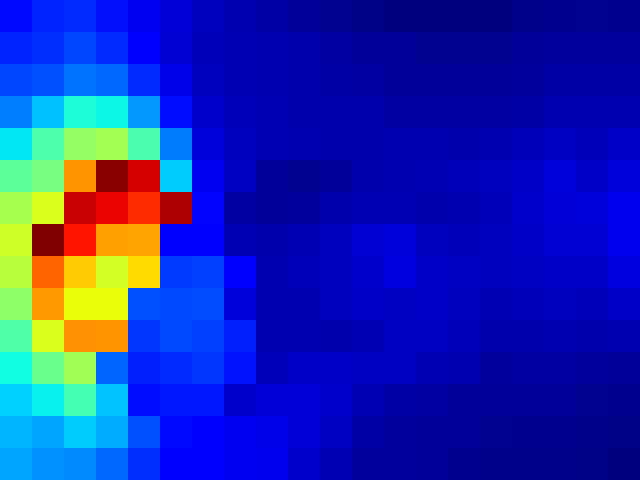}};
\node (sa2) at (4, 0.93) {\includegraphics[width=0.2\linewidth]{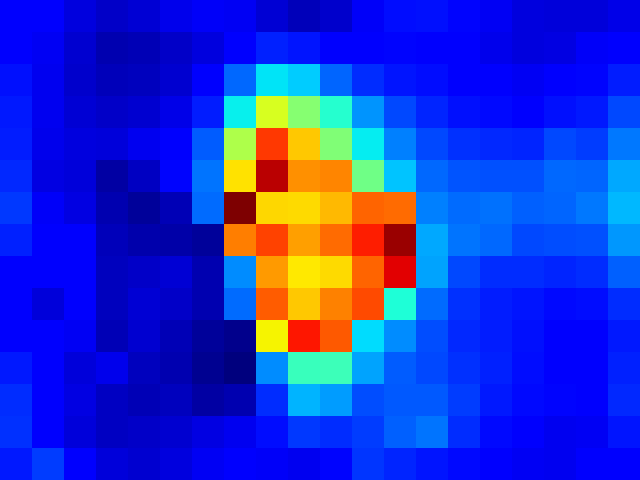}};
\node (sa3) at (4, -0.93) {\includegraphics[width=0.2\linewidth]{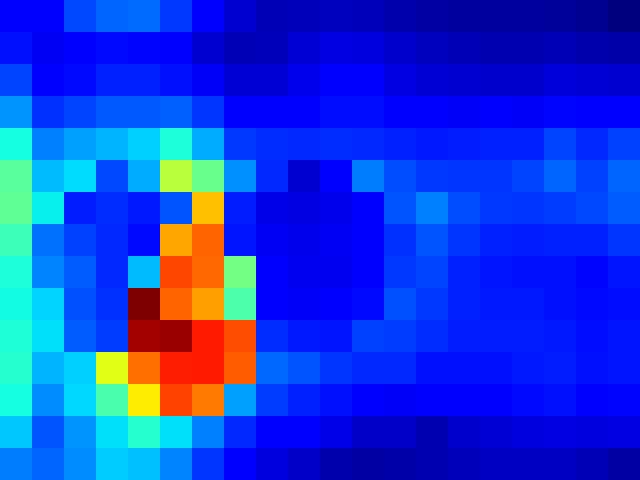}};

\coordinate (sa1_pixel) at (-1.65, -0.3);
\coordinate (sa2_pixel) at (-0.1, -0.3);
\coordinate (sa3_pixel) at (-1.2, -0.3);

\draw [very thick, ->, green]    (sa1_pixel) to[out=90, in=0] (sa1.east);
\draw [very thick, ->, green]    (sa2_pixel) to[out=90, in=180] (sa2.west);
\draw [very thick, ->, green]    (sa3_pixel) to[out=-90, in=180] (sa3.west);

\coordinate (h1) at (-2.45, -0.3);
\coordinate (h2) at (2.45, -0.3);

\draw [thick, -, dashed, blue] (h1) --(h2);

\end{tikzpicture}